\documentclass[letterpaper, 10 pt, conference]{ieeeconf} 

\IEEEoverridecommandlockouts
\usepackage{adjustbox}
\usepackage{svg}
\usepackage{amsmath}
\usepackage{algorithm}
\usepackage{algpseudocode}
\usepackage{booktabs} 
\usepackage{array}
\usepackage{balance}
\usepackage{graphicx}
\usepackage{authblk}
\usepackage{tikz}
\usepackage{pgfplots}
\pgfplotsset{compat=1.17}
\usepackage[hyphens]{url}
\newcommand{\href}[2]{#2}
\usepackage{cleveref}
\usepackage{cite}
\usepackage{dirtytalk}
\usepackage{dblfloatfix}
\usepackage{comment}
\usepackage{tabularx}
\usepackage{makecell}
\usepackage[table]{xcolor}
\usepackage[commandnameprefix=always]{changes}
\usepackage{listings}
\usepackage[acronym]{glossaries}

\newacronym{LLM}{LLM}{large language model}
\newacronym{HRI}{HRI}{human-robot interaction}
\newacronym{SHR}{SHR}{Social Humanoid Robot}

\newcommand{\identitycomponent}{\textsc{identity}}
\newcommand{\capabilitycomponent}{\textsc{capability boundary}}
\newcommand{\transparencycomponent}{\textsc{transparency}}
\newcommand{\taskcomponent}{\textsc{task}}
\newcommand{\failurecomponent}{\textsc{expectation/failure protocol}}
\newcommand{\privacycomponent}{\textsc{privacy}}
\newcommand{\adaptationcomponent}{\textsc{user-adaptation}}
\newcommand{\ethicalcomponent}{\textsc{ethical red lines}}

\usepackage{etoolbox}

\makeatletter
\patchcmd{\thebibliography}{\footnotesize}{\scriptsize}{}{}

\makeatother
\usepackage{cleveref}

\title{\bf Why did My Robot Just Change Personality?\\\adjustbox{max width=\textwidth}{Prompting Guidelines for a Grounded Robot Persona in LLM-Based HRI}}

\author{
\begin{minipage}{0.96\textwidth}
\centering
Ashita Ashok$^{1}$ \quad
Franziska Babel$^{2}$ \quad
Patrick Holthaus$^{3}$ \quad
Rucha Khot$^{4}$ \quad
Karla Bransky$^{5}$\\
Fethiye Irmak Dogan$^{6}$ \quad
Karsten Berns$^{1}$ \quad
Silvia Rossi$^{7}$ \quad
Minha Lee$^{4}$ \quad
Guy Laban$^{8,9,10,*}$\\

\small
$^{1}$Robotics Research Lab, RPTU Kaiserslautern, Kaiserslautern, Germany\\
$^{2}$Human-Centered Systems, Link{\"o}ping University, Link{\"o}ping, Sweden\\
$^{3}$Robotics Research Group, University of Hertfordshire, Hatfield, UK\\
$^{4}$Industrial Design, Eindhoven University of Technology, Eindhoven, Netherlands\\
$^{5}$Gameflow Lab, Australian National University, Canberra, Australia\\
$^{6}$Affective Intelligence and Robotics Laboratory, University of Cambridge, Cambridge, UK\\
$^{7}$Intelligent Robotics and Advanced Cognitive System Projects Laboratory, University of Naples Federico II, Naples, Italy\\
$^{8}$Department of Industrial Engineering and Management, Ben-Gurion University of the Negev, Beer Sheva, Israel\\
$^{9}$School of Brain Sciences and Cognition, Ben-Gurion University of the Negev, Beer Sheva, Israel\\
$^{10}$The Azrieli National Center for Autism and Neurodevelopment Research, Beer Sheva, Israel\\

\footnotesize $^{*}$Corresponding author: \href{mailto:laban@bgu.ac.il}{\texttt{laban@bgu.ac.il}}
\end{minipage}
}

\begin{document}

\maketitle
% \thispagestyle{empty}
% \pagestyle{empty}

%%%%%%%%%%%%%%%%%%%%%%%%%%%%%%%%%%%%%%%%%%%%%%%%%%%%%%%%%%%%%%%%%%%%%%%%%%%%%%%%
\begin{abstract}
Large language models (LLMs) are increasingly used for verbal interaction in social robots, yet prompt design in human-robot interaction (HRI) remains underspecified. As a result, robots may present hallucinated capabilities, unclear behavioural boundaries, and misleading personas. This paper develops a framework for prompt design in LLM-based robots and introduces a structured prompt template comprising eight functional components through which robot behaviour can be specified, bounded, and adapted. The framework is grounded in a review of prior LLM-based HRI work and complemented by survey and discussion data from HRI experts gathered at the Robo-Identity workshop at IEEE RO-MAN 2025 ($N=27$). The qualitative findings highlight limited legibility of robot personality, the need for user adaptation, and strong ethical concerns about safety, deception, and governance. Based on these findings, we present prompting guidelines accompanied by proof-of-concept template as a structured design and reporting aid for HRI research. We argue that prompt design should be treated as a socio-technical problem rather than a minor implementation detail, requiring explicit capability boundaries, transparent behavioural assumptions, and context-sensitive safeguards to support reliable and interpretable HRI.
\end{abstract}

% \begin{IEEEkeywords}
% human-robot interaction, social robots
% \end{IEEEkeywords}

%%%%%%%%%%%%%%%%%%%%%%%%%%%%%%%%%%%%%%%%%%%%%%%%%%%%%%%%%%%%%%%%%%%%%%%%%%%%%%%%
\vspace{-3mm}
\section{Introduction} %please note original writing moved to original.tex 
% 346 words 
%7 unique citations

Social interactive robots, particularly those with a human-like appearance, increasingly rely on verbal interaction powered by \glspl{LLM} \cite{Laban2024SharingFeel,kim2024survey}. These models can support natural language conversation and provide access to broad world knowledge \cite{Laban2024SharingFeel}, but they are also prone to hallucination \cite{Ji2023halluc} and remain fundamentally \textit{disembodied} from the robot's hardware, environment, and user expression. In \gls{HRI}, this gap can produce fluent but misleading interaction, for example, when a robot implies knowledge, perception, or capabilities that it does not actually possess \cite{skantze2025applying}. Grounding robot behaviour in its actual capabilities and interaction context is therefore critical for calibrating user expectations and trust \cite{kraus2020more}. As \glspl{LLM} increasingly shape robot communication, prompt design becomes a central \gls{HRI} problem rather than a minor implementation detail: before interaction begins, \gls{LLM}-based robots must be configured through structured prompt components that define the robot's identity, capabilities, disclosures, and behaviour under uncertainty.
    
\begin{figure}
    \centering
\includegraphics[width=\linewidth]{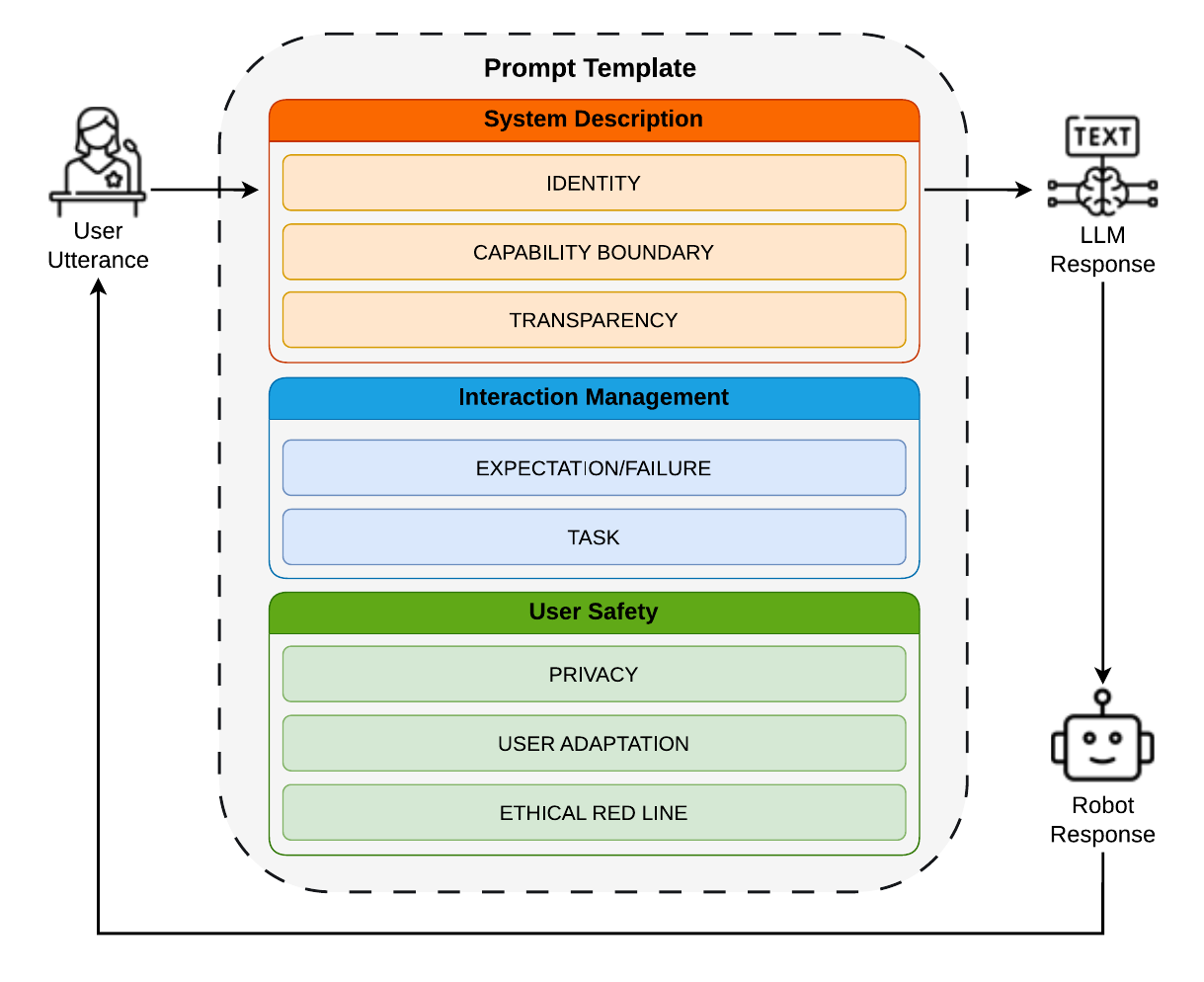}
\caption{A conceptual framework for prompt design in LLM-powered robots.}
    \label{fig:teaser}
    \vspace{-5mm}
\end{figure}

Accordingly, this paper aims to clarify \textit{what} should be specified in such prompts and \textit{why}. We pursue this goal in two complementary steps. First, we draw on broader \gls{HRI} scholarship to develop a theoretical framework for understanding how robot identity, behaviour, expectations, transparency, and behavioural boundaries are currently constructed and interpreted in embodied interaction, and how these may shape prompt design as a central mechanism of robot behaviour and identity. We then use this framework as an analytic lens to survey selected recent \gls{LLM}-based \gls{HRI} studies and identify the core prompt components through which these aspects are specified, bounded, or left implicit in practice, leading to a recommended prompt template for \gls{HRI}. Second, we complement this framework and literature-derived template with an empirical layer based on an expert survey and discussion from three HRI expert groups during the workshop \href{https://rrlab.cs.rptu.de/en/conferences/ro-man-2025}{\say{Robo-Identity
: Methods \& Theories in Personification of Social Robots}} at RO-MAN 2025 with \gls{HRI} researchers and practitioners. Together, these two steps allow us to address three linked challenges in \gls{LLM}-based \gls{HRI}: supporting a consistent and interpretable robot identity and behaviour s\cite{Laity2025,bransky_identity_2026}, reducing unnecessary deception and misleading interaction \cite{babel2026anthropomorphic,esposito2025deception}, and limiting hallucinations and capability mismatches in embodied settings \cite{atuhurra2024leveraging,skantze2025applying}. The outcome of this work is a structured prompt template comprising eight functional components (see Fig~\ref{fig:teaser}): \identitycomponent{}, \capabilitycomponent{}, \transparencycomponent{}, \taskcomponent{}, \failurecomponent{}, \privacycomponent{}, \adaptationcomponent{}, and \ethicalcomponent{}. We present this template %not as a fixed recipe for all robots and contexts, but 
as a structured design and reporting aid for \gls{HRI} research rather than a fixed recipe for all robots and contexts. In this way, the paper contributes a  %practical 
framework for defining bounded artificial robot personas that align with system capabilities, make prompt assumptions explicit, and support more transparent, comparable, and reproducible \gls{LLM}-based \gls{HRI}. %At the same time, the paper contributes a theoretical framework for understanding the scope and implications of prompting robot identity and behaviour.

\section{Theoretical Framework}\label{sec:ms}

Prompt design can be understood as a central mechanism through which robot behaviour is configured for interaction. Rather than treating behaviour as an emergent product of language generation alone, this perspective highlights that prompts shape how a robot is presented, what it is expected to do, and how its conduct is constrained in social interaction. Drawing on prior HRI scholarship on robot identity, expectation formation, transparency, and interaction failure, we use this framework to examine how prompt design structures artificial personas and their boundaries in embodied HRI.

%Based on the studies reviewed in Section~\ref{sec:rw} and relevant literature, we have developed a theoretical framework for prompt design in LLM-based robots. The framework treats robot behaviour not as an emergent product of language generation alone, but as something that must be explicitly shaped through prompt-level specifications of identity, capability boundaries, transparency, task, and user-facing safeguards. In this way, the framework connects prior findings on robot identity, expectation formation, transparency, and interaction failure to a structured account of how artificial personas should be defined and constrained in HRI.
%In the following, we detail selected initial prompt components of this framework as identified in our analysis of \gls{HRI}-specific literature, which will then be refined via our qualitative assessment and common failures in \gls{LLM}-powered robots.

\paragraph{Identity} %please note original writing moved to original.tex 
Identity in \gls{HRI} is not peripheral but central, emerging through design strategies, social dynamics, and users' own constructions of robot personas. Many aspects contribute to perceived identity, including embodiment and behavioural cues~\cite{bransky_identity_2026}. Users rely on stereotype projection based on appearance (e.g., gender, ethnicity, or racial bias)~\cite{seaborn2025social}, while hybrid embodiments further shape perception~\cite{bransky_identity_2026}, for example, when additional interfaces increase perceived competence~\cite{lee2023here}. Research in Robo-Identity has further highlighted that identity should be understood as a socially constructed and evolving phenomenon~\cite{lee2021robo,laban2022robo,seaborn2022identified,khot2024robo}, shaped over time through interaction and context~\cite{seaborn2025social}. Prior work on artificial identity design also outlines key principles for specifying identity attributes, including role, persona, embodiment, and functional constraints, but does not operationalise these for \gls{LLM}-based systems~\cite{bransky_identity_2026}.

Operationalising identity in prompts therefore requires explicit specification of a coherent set of characteristics. This includes the robot’s role and persona (e.g., communication style, expressiveness, and behavioural tone) \cite{bransky_identity_2026}, as well as constraints derived from embodiment and system capabilities. Prior work highlights that identity is closely tied to embodiment, system design, and user perception \cite{bransky2024mind, bransky_identity_2026}, requiring designers to define how an agent is presented and understood in context. Therefore, robot identity should be treated as a designed yet evolving construct that maintains behavioural consistency while remaining aligned with embodiment, capabilities, and interaction context.

\paragraph{Transparency} %please note original writing moved to original.tex 
Transparency by Design in \gls{HRI} points to behavioural strategies that make a robot’s intentions, decisions, and internal states intelligible to human users, thereby supporting safety, predictability, trust, and effective collaboration. Transparency is a functional requirement in contexts where humans must rapidly interpret robot behaviour and coordinate with autonomous systems. Accordingly, transparency involves making robot behaviour and internal states interpretable and legible at a communicative level~\cite{Holthaus2023} using signals like social gaze~\cite{vigni2022exploring}. It includes not only how robots signal communicative intent, but also how they convey confidence, uncertainty, and progress during interaction, for example, via inner speech and emotional expression~\cite{angelopoulos2022you,angelopoulos2023unveiling}. Such cues help users form a more accurate understanding of what the robot is doing and why. Notably, the intention to be transparent is not the same as being transparent in an interaction. For instance, roboticists may aim to transparently convey information across different affordances like a robot verbally or visually (e.g., using a screen) describing what it is doing, but human interactants may still face high informational load, and thus interactional transparency is not delivered as intended~\cite{lee2023here}.

The use of social and emotional cues introduces the ethically complex issue of deception in \gls{HRI}. Although some robot behaviours and designs may, in some cases, increase compliance, likeability, or perceived agency \cite{Laban2021TellSpeech,Laban2024BuildingTime,Laban2025CopingCaregivers,Laban2025WhatTime}, they also risk undermining trust, particularly when users interpret the deception as deliberate~\cite{rossi2024way}. Systematic analysis further highlights the broader implications of deception in HRI and the need for safeguards~\cite{esposito2025deception}. One mitigation strategy is to incorporate behavioural cues that implicitly signal uncertainty or deception, thereby preserving a degree of transparency~\cite{esposito2025roboleaks}. Overall, transparency in \gls{HRI} should be understood as a multidimensional construct encompassing legibility, predictability, and the explicit communication of system limits, complicated by if and how multimodal communication, e.g., speech and GUI presentation, actually helps or hinders transparency in interactions.

\paragraph{Explanations and Expectations} %please note original writing moved to original.tex
Shaping robot behaviour in HRI requires acknowledging that people perceive robots holistically~\cite{torre2021voice}; accordingly, explanations and expectations should be treated as a coupled design problem rather than as separate concerns. Expectations define what users believe a robot can or should do, while explanations help establish, calibrate, and revise those beliefs over time. From this perspective, explanations are not merely post-hoc accounts of robot decisions; they are part of the mechanism through which robots communicate capability, limitation, and social role~\cite{yadollahi2025expectations, dogan2026designing}. This is especially important in socially situated interaction, where behaviour must remain intelligible, socially appropriate, and aligned with human preferences~\cite{dougan2025grace}. 

In this context, explanations are closely tied to expectation formation and adjustment. In socially situated environments, common-sense reasoning alone is often insufficient, particularly when appropriateness depends on context or when users hold differing expectations. Prior work shows that incorporating human explanations into action generation can improve alignment with social norms and produce more interpretable robot behaviour~\cite{dougan2025grace}. At the same time, expectations shape how explanations are perceived, particularly in situations of failure and recovery. Evidence from priming and failure-recovery studies suggests that expectations can significantly influence perceived robot competence, and that explanations are more effective when they support calibrated expectations~\cite{yadollahi2025expectations}. These findings highlight the need to design explanations that not only account for behaviour, but also embodiment and environment to meet people's expectations during an interaction.

\begin{comment}

\begin{figure*}[h!]
    \centering
\includegraphics[width=.9\textwidth]{figures/Barchart_paperplaza_pdf14.pdf}
\caption{Bar chart showing the agreement or disagreement to the 12 statements about \glspl{LLM} and Personas presented to the 27 workshop participants in three groups. Topics: 1) Personas as Design Constructs (items 1-3), 2) Anthropomorphism and Deception (items 4-5), 3) AI-Driven Personas (6-9), 4) Deception and Ethics (items 10-13)}
    \label{fig:diagram}
\end{figure*}
\end{comment}

%\paragraph{Common Pitfalls in Prompt Design}
%Prior work reveals recurring failures in LLM-based robot deployments when key prompt components are underspecified. These include hallucinated capabilities (\capabilitycomponent), ambiguous behaviour (\taskcomponent), undisclosed use of user data (\privacycomponent), mismatches with user expectations (\adaptationcomponent), and harmful outputs (\ethicalcomponent), motivating their inclusion of the remaining prompt components.

\section{Prompt Design in LLM-Based HRI}
\label{sec:rw}

\begin{comment}
    
The deployment of \glspl{LLM} in \glspl{SHR} has accelerated following the release of GPT models and accessible APIs in 2022, enabling conversational robot behaviours such as dialogue generation, emotion expression, and interaction management \cite{williams2024scarecrows}. This rapid deployment has prompted discussions in the HRI community on methodological and reporting practices for integrating \glspl{LLM} into robot interaction pipelines \cite{matuszek2026reporting}. In such systems, prompts act as the primary mechanism for conditioning LLM behaviour, shaping how the robot presents itself, responds to users, and manages dialogue flow \cite{kim2024survey}.
\end{comment}

Prior work on LLM-based HRI has introduced a growing range of robot applications, but has been far less explicit about how robot behaviour is actually specified in prompts. To situate our framework, we surveyed selected recent LLM-based HRI studies from the past three years that explicitly reported prompt design (see Table~\ref{tab:rw_llms}), with attention to which aspects of robot behaviour were specified, constrained, or left implicit. Given the still limited and rapidly evolving nature of this literature, we used a selected sample of representative studies to map emerging prompting practices rather than to provide an exhaustive systematic review. Our analysis was guided by one central question: which aspects of robot behaviour are being specified, bounded, or left implicit at the prompt level? This question was informed by the theoretical framework (see Section~\ref{sec:ms}), which treats robot persona and behaviour as the mechanism through which a robot’s identity, behavioural boundaries, transparency, interactional purpose, and user-facing safeguards are defined. Accordingly, for each paper, we examined how the prompt specified the robot’s role and persona, interaction goal, handling of uncertainty or failure, and whether it made explicit any capability limits, disclosures, privacy assumptions, user-adaptation strategies, or ethical constraints. %On this basis, we derived an initial eight-component template.

 %we selected and reviewed recent (from the past 3 years) representative studies that report prompt design in HRI studies (see Table~\ref{tab:rw_llms}), and examined which aspects of robot behaviour are defined, constrained, or left implicit. 
 Most reviewed works define the robot identity within the prompt (90\%) and specify the interaction task (60\%). Other prompt elements appear less consistently. Expectation or failure-handling strategies are present in about half of the prompts (50\%), while explicit user adaptation is rare (30\%). Privacy-related instructions appear only sporadically (20\%), despite several systems storing conversation histories or maintaining persistent user profiles. Explicit ethical constraints are reported in a limited subset of prompts (30\%). Notably, none of the reviewed prompts specifies the robot's embodiment or system capability boundaries (e.g., sensing limitations, mobility constraints, or knowledge access), nor do they disclose the use of LLM to the user. Consequently, both the capability boundary and transparency components remain largely unaddressed. Overall, current prompt designs in LLM-powered HRI primarily establish robot identity and task, while other aspects important for predictable and responsible interaction remain inconsistently specified. %Table~\ref{tab:promptcomponents} highlights representative prompt excerpts illustrating these components across the reviewed works.

%Table~\ref{tab:rw_llms} summarises representative works deploying LLM-powered SHRs, selected for explicitly reporting LLM prompts, enabling a structured comparison of prompt design in HRI.

\begin{table}[h!]
\centering
\vspace{1mm}
\caption{\small Comparison of relevant works on LLM-powered SHRs}
\vspace{-1mm}
\label{tab:rw_llms}
\begin{tabular}{p{1.9cm}ccccc}
\hline
\textbf{\makecell{Reference\\(Sample)}} & \textbf{Model} & \textbf{\makecell{Social\\Robot}} & \textbf{Use Case}\\
\hline
\makecell[l]{Billing et al.\cite{billing2023language}\\(N=N/A)} & \makecell{OpenAI\\GPT-3} & \makecell{Pepper\\\& Nao} & \makecell{Robot\\dialogue\\generation}\\
\hline
\makecell[l]{Kim et al.\cite{kim2024understanding}\\(N=32)} & \makecell{OpenAI\\GPT-3.5} &
\makecell{Pepper} & \makecell{LLM capability\\evaluation}\\
\hline
\makecell[l]{Verhelst et al.\cite{verhelst2024adaptive}\\(N=21)} & \makecell{OpenAI\\GPT-3.5} & Furhat & \makecell{Language\\tutoring}\\
\hline
\makecell[l]{Addlesee et al.\cite{addlesee2024multi}\\(N=N/A)} & Vicuna-13b-v1.5 & ARI & \makecell{Multi-party\\HRI}\\
\hline
\makecell[l]{Kim et al.\cite{kim2024child}\\(N=24)} & \makecell{OpenAI\\GPT-4o-mini\\(fine-tuned)} & piBo & \makecell{Child-centric\\intention-aware\\HRI}\\
\hline
\makecell[l]{Skantze et al.\cite{skantze2025applying}\\(N=39)} & \makecell{OpenAI\\TurnGPT\\(fine-tuned)} & Furhat & \makecell{Turn-taking\\modelling}\\
\hline
\makecell[l]{Pinto et al.\cite{pinto2025designing}\\(N=50)} & \makecell{OpenAI\\GPT-3.5} & Pepper & \makecell{Memory-aware\\HRI with\\elderly}\\
\hline
\makecell[l]{Mauliana et al.\cite{mauliana2025exploring}\\(N=13)} & \makecell{GoogleAI\\Flan-T5-Large\\(fine-tuned)} & Ameca & \makecell{Memory-aware\\multi-session HRI\\ with students}\\
\hline
\makecell[l]{Sievers et al.\cite{sievers2025using}\\(N=N/A)} & \makecell{OpenAI\\GPT-4o} & Pepper & \makecell{Knowledge\\retrieval\\ assistance}\\
\hline
\makecell[l]{Laban et al.\cite{laban2026robot}\\(N=21)} & \makecell{OpenAI\\GPT-3.5\\} & QTrobot & \makecell{Emotion\\regulation\\intervention\\}\\
\hline
\end{tabular}
\vspace{-4mm}
\end{table}

%To analyse prompt design in prior work, prompt content is operationalised into eight functional components. These components were defined through iterative expert-driven synthesis among the authors, informed by prior literature, workshop findings, and practical experience with \gls{LLM} deployment. They reflect recurring dimensions observed across prior work rather than an arbitrary taxonomy.
Based on the reviewed studies, we identified eight prompt components that HRI researchers should report when documenting LLM-based robotic systems. Explicitly describing these components helps make clear how robot behaviour is defined, bounded, and adapted, while supporting transparency, comparability, and reproducibility in the field.

\begin{itemize}
\item \identitycomponent: Defines the robot’s persona and social role assumed by the language model.
\item \capabilitycomponent: Specifies the robot’s perceptual inputs, embodiment constraints, and system limitations (e.g., sensing, memory, mobility, internet access).
\item \transparencycomponent: Specifies whether and how the robot system discloses its artificial nature or operational limitations to the user.
\item \taskcomponent: Defines the primary conversational objective the robot should accomplish.
\item \failurecomponent: Specifies how the robot handles uncertainty, missing information, or interaction breakdowns.
\item \privacycomponent: Specifies constraints on the collection, storage, and use of user data by robot system.
\item \adaptationcomponent: Defines how the robot adapts communication to the target user group (e.g., age, language, cognitive ability).
\item \ethicalcomponent: Specifies constraints preventing harmful, biased, unsafe, or illegal robot responses.
\end{itemize}

%\section{Sample Prompt and Demo}\label{sec:sp}

To illustrate how the identified prompt components can be operationalised in practice, 
we provide two \textbf{complete prompt examples} using all eight modular design components 
as an OSF project\footnote{https://osf.io/qrnzp/files/9xbpr} (see Section \textit{Sample Prompt}). An \textbf{interactive demo} is available on Hugging Face\footnote{https://huggingface.co/spaces/aashok/LLM-PSR}, 
and the corresponding \textbf{implementation} is released on GitHub\footnote{https://github.com/aashok94/llm\_powered\_social\_robot}.

% \begin{table}[h]
% \centering
% \caption{Representative prompt excerpts for identified prompt components.}
% \label{tab:promptcomponents}
% \begin{tabular}{p{2cm}p{5.6cm}}
% \hline
% \textbf{Component} & \textbf{Excerpt} \\
% \hline

% Identity 
% & ``I want you to act as the robot Pepper. We are visiting the senior fair at Arena in Skövde.'' \cite{billing2023language} \\
% \hline
% Capability Boundary
% & Not explicitly reported in the reviewed works. \\
% \hline
% Transparency
% & Not explicitly reported in the reviewed works. \\
% \hline
% Task
% & ``Generate a short one-sentence description in Spanish of a picture that contains a \{word\}.'' \cite{verhelst2024adaptive} \\
% \hline
% Expectation/Failure
% & ``If content is not found, you tell the traveller that you don't know the answer and do not make up an answer.'' \cite{sievers2025using} \\
% \hline
% Privacy
% & ``Here’s what you know about them: user\_info. Previous conversations included: past\_interactions.'' \cite{pinto2025designing} \\
% \hline
% User Adaptation
% & ``Your primary users are children aged 5-7 years. Provide play and conversation activities.'' \cite{kim2024child} \\
% \hline
% Ethical Red Line
% & ``Your answers should not include harmful, unethical, racist, sexist, toxic, dangerous, or illegal content.'' \cite{mauliana2025exploring} \\

% \hline
% \end{tabular}
% \end{table}

\section{Qualitative Assessment}\label{sec:su}

Building on the theoretical framework, we introduce a complementary empirical layer based on a participatory expert survey and discussion conducted at the Robo-Identity workshop. The purpose of this data collection is not to derive the framework inductively, but to empirically examine how HRI researchers and practitioners interpret its core concerns, including persona legibility, continuity over time, safety boundaries, and context-sensitive governance. As a complementary empirical layer, this study draws on data collected during a workshop at RO-MAN 2025\footnote{https://rrlab.cs.rptu.de/en/conferences/ro-man-2025}, including a survey and three focus groups. The workshop examined how artificial personas in social robots are constructed, interpreted, and discussed within the context of recent advances in \glspl{LLM} and generative AI.
%274 words
%11 unique citations
%This study draws on data collected during a workshop conducted at \href{https://rrlab.cs.rptu.de/en/conferences/ro-man-2025}{RO-MAN 2025}. The workshop aimed to examine how artificial personas in social robots are constructed and interpreted, particularly in light of recent advances in \glspl{LLM} and generative AI. %While these technologies enable the rapid creation of robot personas through prompting, they also raise important ethical questions related to deception, persuasion, and user trust. The workshop therefore focused on discussing methodological approaches, theoretical perspectives, and ethical considerations in the design of social robots with AI-generated personas. 
%The workshop consisted of keynotes and presentations and finished with a group discussion structured by a survey which was filled in by the group leader. 
This participatory study was approved by the Ethics Committee of the Department of Social Sciences, RPTU Kaiserslautern (Approval No. SoWi/69). Following the presentations in the workshop, participants formed three discussion groups. Two workshop organizers joined the groups to facilitate the discussions, while each group collaboratively completed a structured survey summarizing their perspectives and discussion outcomes.

\begin{figure*}[t]
    \centering
\includegraphics[width=.92\textwidth]{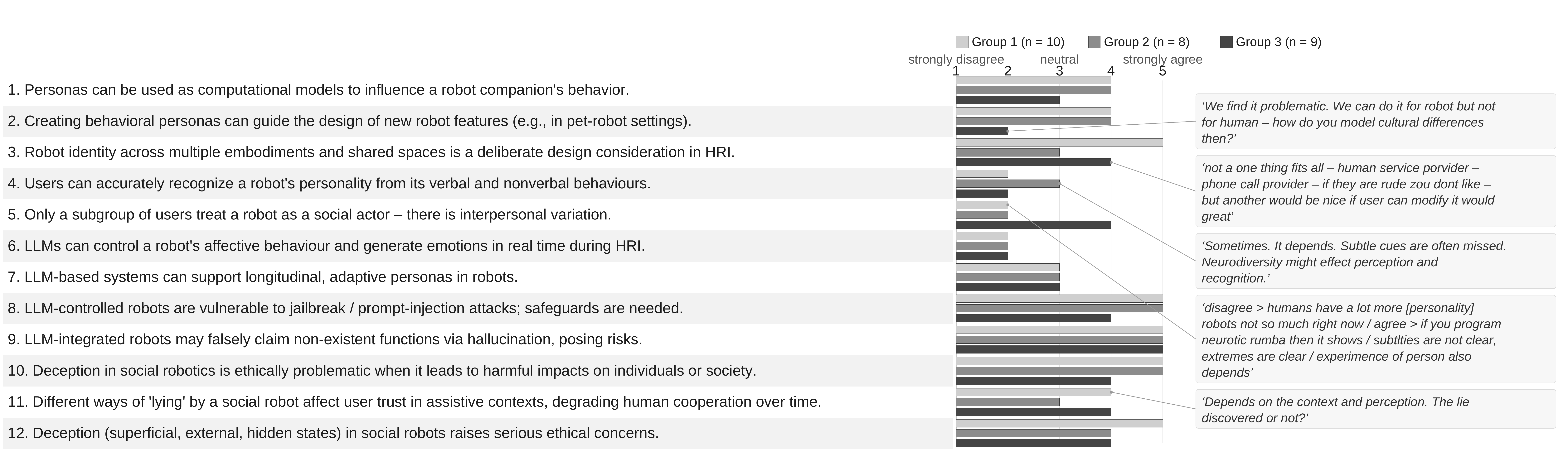}
\caption{\footnotesize Bar chart showing the agreement or disagreement to the 12 statements about \glspl{LLM} and Personas presented to the 27 workshop participants in three groups. Topics: 1) Personas as Design Constructs (items 1-3); 2) Anthropomorphism and Deception (items 4-5); 3) AI-Driven Personas (6-9); 4) Deception and Ethics (items 10-13).}
    \label{fig:diagram}
    \vspace{-3mm}
\end{figure*}

%\subsection{Participants}
All participants were experts in HRI (most $>$ 5-9 years of experience), of different cultures and locations (Asia, Europe, US, Oceania), as well as backgrounds extending HRI (computer science, psychology, ethics, AI, social sciences). Participants were divided into three groups (Group~1: $n=10$, Group~2: $n=8$, Group~3: $n=9$), resulting in a total sample of $N=27$ participants. Each group collaboratively completed a single 12-item survey with open-ended statements designed to elicit group discussion on different aspects of robotic persona and to capture the main outcomes of the group deliberation. In addition, the discussions included two concluding free-response questions (persona adaptation for different user groups; ethical red line in persona design). Each statement included a Likert scale  (1 = strongly disagree, 5 = strongly agree) and the items were designed to capture participants’ perspectives on the design and interpretation of robot personas in AI-enabled social robots. The full questionnaire, including item sources and supporting literature, is available on OSF\footnote{https://osf.io/qrnzp/files/9xbpr}. Thematic analysis was conducted at a semantic level on the data collected by two authors independently, focusing on how participants reasoned about robot persona and identity, their recognizability, continuity over time, and ethical implications.

%\subsection{Item Generation}
%The survey combined 12 Likert statements (1 = strongly disagree, 5 = strongly agree) with accompanying open text fields, as well as two concluding free-response questions (persona adaptation for different user groups; ethical red line in persona design).
%The survey questions were designed to capture participants’ perspectives on the design and interpretation of robot personas in AI-enabled social robots. 
%The full questionnaire, including item sources and supporting literature, is available on OSF\footnote{\url{https://osf.io/qrnzp/overview?view_only=121d0f7caace404ea53e10aa92563166}}, Section \textit{Questionnaire}; individual items are also stated in Fig.~\ref{fig:diagram}.

%The items were derived from prior work on robot personas and behavioural design models \cite{duque2013different,dos2014behavioral,staffa2024influence}, anthropomorphism and user perception of robot personality \cite{spitale2023robotic,fischer2011interpersonal}, recent developments in AI-enabled robotic systems and LLM-driven behaviour generation \cite{mishra2023real,robey2025jailbreaking,laban2024sharing}, and ethical considerations surrounding deception and trust in social robotics \cite{sharkey2021we,rossi2024way}. 

\section{Results of Thematic Analysis}\label{sec:th}

    \begin{table}[t]
\caption{\small Themes from the qualitative analysis.} %Means are reported descriptively across the three group submissions.}
\label{tab:joint_display}
\centering
\scriptsize
\setlength{\tabcolsep}{3pt}
\renewcommand{\arraystretch}{1.05}
\begin{tabular}{>{\raggedright\arraybackslash}p{1.3cm}
                >{\raggedright\arraybackslash}p{3.2cm}
                >{\raggedright\arraybackslash}p{3.1cm}}
\hline
\textbf{Theme} & \textbf{Qualitative interpretation} & \textbf{Illustrative quote} \\
\hline

\textbf{Persona as a Useful but Contested Design Abstraction} &
Participants saw persona as technically possible, but questioned whether it is always the right or sufficiently well-defined design direction. &
\textit{``Yes --- it can, the question, should it\ldots\ we should discuss whether it is the right approach.''} (Group 2) \\\midrule

\textbf{Robot Personality Is Not Necessarily Legible} &
Participants doubted that users reliably infer stable personality from robot behavior, especially when cues are subtle and user interpretation varies. &
\textit{``Subtle cues are often missed.''} (Group 2) \newline
\textit{``subtlties are not clear, extremes are clear.''} (Group 3) \\\midrule

\textbf{Identity Over Time Was Framed as a Memory Problem} &
Longitudinal persona was framed as requiring memory, temporal ordering, and coherent state across interactions, rather than prompt style alone. &
\textit{``It can support, but would be limited in memory long term due to limited context window.''} (Group 2) \\\midrule

\textbf{Safety and Ethics Structured the Discussion} &
Participants evaluated persona and identity primarily through risks of hallucination, adversarial prompting, and deceptive or unsafe outputs. &
%\textit{``duh\ldots\ internal safeguard.''} (Group 3) \newline
\textit{``Depends on context, deception is part
of robots design.''} (Group 2) \newline
\textit{``people have died from llm suggestions.''} (Group 3) \\\midrule

\textbf{Context-Sensitive and Vulnerability-Aware Design} &
Participants rejected one-size-fits-all personas and emphasised tailoring, transparency, privacy boundaries, and greater caution in vulnerable contexts. &
\textit{``not a one thing fits all.''} (Group 3) \newline
\textit{``Not being transparent\ldots\ using the persona of a real person\ldots''} (Group 2) \\

\hline
\end{tabular}
\vspace{-1em}
\end{table}

%please note original writing moved to original.tex 
%892 words
%Each group collaboratively completed one survey reflecting their discussion outcomes; Group~2 did not provide open-ended responses. Open-text answers were obtained from two groups (25 text segments in total). Given this structure, 
Results are presented as a qualitative-led analysis, with Likert responses used descriptively to indicate convergence and divergence across groups. Groups were mostly in agreement about most statements, with the exceptions of items 2, 3, 4, 5 and 11 (see Fig.~\ref{fig:diagram}).

\subsection{Persona as a Useful but Contested Design Abstraction}
Consistent across the responses was that participants did not reject the idea of robot persona outright, neither did they treat it as desirable. Instead, persona was positioned as a potentially useful design abstraction whose legitimacy depends on how it is defined and what work it is expected to do. Group 2 in response to item 1 acknowledged the technical feasibility of generating companion-like behaviour, but reframed the issue as normative: \textit{``Yes --- it can, the question, should it [...] we should discuss whether it is the right approach.''} Group 3 expressed a similar hesitation questioning how such an approach could account for contextual variation such as culture. This tension was also visible in the ratings of items 1-3 (see Fig \ref{fig:diagram}).
%Participants generally regarded robot persona as technically feasible, but not as an automatically justified or sufficiently well-defined design direction. Rather than rejecting persona itself, they questioned when it should be used and on what basis. Group 2 captured this tension directly: \textit{``Yes --- it can, the question, should it\ldots\ we should discuss whether it is the right approach.''} Group 3 raised a related concern about reductionism, describing the categorisation of persona into discrete parts as an \textit{``unclear item''} and questioning whether such an approach could account for broader contextual influences such as culture. %This tension was also visible in the ratings of items 1--3. Participants moderately agreed that personas can shape robot behaviour (\(M = 3.67\)), but were more mixed on whether personas should guide the design of new features (\(M = 3.33\)). Overall, persona was treated as possible and potentially useful, but with .

\subsection{Robot Personality Is Not Necessarily Legible}
A second pattern concerned the difficulty of making robot personality reliably legible. Participants suggested that subtle behavioural cues are unlikely to translate into stable interpretations across users. Group 2 stated that \textit{``Subtle cues are often missed,''} adding that \textit{``Neurodiversity might affect perception and recognition.''} Group 3 further clarified that personality may only be recognisable under exaggerated conditions (e.g., neurotic Roomba) \textit{``subtleties are not clear, extremes are clear,''} and that interpretation depends on user experience. This was also reflected in the ratings. Groups expressed low agreement that users can accurately recognise a robot’s personality, and disagreed that LLMs can generate emotions in real time.

\subsection{Identity Over Time Was Framed as a Memory Problem} 
When discussing continuity, participants framed identity as dependent on memory and system architecture rather than prompt design alone. Group 2 noted that longitudinal persona is limited by context, while Group 3 emphasised that LLMs alone cannot sustain continuity without memory of prior interactions and temporal sequencing. Participants did not reject adaptive identity, but conditioned it on the presence of mechanisms for retaining and organising interaction history.

\subsection{Safety and Ethics Structured the Discussion}
Participants consistently framed persona and identity through risks related to hallucination, manipulation, and misleading representations. Group 3 described safeguards as necessary (\textit{``internal safeguard''}) and referenced real-world harms (\textit{``people have died from llm suggestions''}). They also highlighted that risk extends beyond physical actions to verbal outputs. Group 2 emphasised that the impact of deception on trust depends on context and whether it is detected. The answers to statement 12, whether deception in social robots raised ethical concerns were interesting as Group 2 stated that it \textit{``Depends on context, deception is part of robots design.''} while Group 3 thought that \textit{``We should be more careful when designing prompts with vulnerable groups; is it lying or not safeguarded; how we check the rights and wrongs.''}.

\subsection{Context-Sensitive and Vulnerability-Aware Design}
Participants argued against one-size-fits-all persona design and emphasised the importance of context, culture, and user vulnerability. Consistent with this, Group 3 also noted that identity across embodiments (i.e., migrating agents) is unlikely to follow a single universal model, suggesting the need for adaptation based on context or user preference. They also emphasised increased caution when designing for vulnerable groups. Participants articulated explicit boundaries. Group 2 identified lack of transparency, use of real-person personas, and harmful behaviour as unacceptable. Group 3 raised privacy concerns and emphasised the need for ethical
oversight. Responses further suggested that persona may evolve over time with the user, reinforcing the need for adaptable and governed interaction design.

\subsection{Overall Interpretation}
Across the dataset, persona was seen as potentially useful, but contingent on legibility, system support for continuity including strong safety and ethical constraints. Agreement was strongest on risks related to hallucination, manipulation, and deception, while claims about stable personality perception and real-time emotional generation were met with scepticism. Disagreement centred on how broadly persona can be generalised as a design tool.

% to dos ashita: 1) check who said LLM-based systems can support longitudinal, adaptive personas in robots. for questionnaire no 7. (DONE) 2) check if lee2022ethics said item 11. (DONE) 3) finalize github code make it public with prompt in json 4) ask help for uploading supplementary to osf your account is MIA or osf dead (DONE) 5) check if submitting 7 or 6 then write discussion. 

\label{sec:dc}
\begin{comment}
    
- how we want the community to use the guidelines
- accept limitations 
- better hri prompting guidelines
- design identity that is not lying, consistent behaviour, not fostering attachment and addiction, not hallucinating
- discuss non-humanoid robot design and the use of LLMs: mismatch of expectations we have based on the embodiment, include "you do not have arms" in the capability prompt: justify why we need the capability prompt
\end{comment}

\section{Discussion and Conclusion}\label{sec:dc}

This paper contributes a structured prompt template for LLM-based robots by combining a review of prior work with a complementary empirical layer from HRI experts. The literature showed that existing systems most often specify \identitycomponent{} and \taskcomponent{}, while components such as \capabilitycomponent{} and \transparencycomponent{} are far less consistently reported. The qualitative findings suggest that this imbalance matters. 
%Participants treated persona not as a stylistic layer alone, but as a socio-technical design problem shaped by embodiment, interpretation, safety, and system boundaries.
HRI experts agreed that if those components are not well defined, robot personas might not be useful as they might not be legible and adapted to the user or even deceptive and harmful to vulnerable groups. 

%A central implication is that current prompting practices need to better define who the robot should be than at defining what it can actually do. 
While the literature-derived template highlights several functional prompt components, the qualitative results show that safety, deception, hallucination, and governance are among the most consequential concerns in practice, which researchers care about. Taken together, these findings suggest that the components least often reported in prior work may be among the most important for responsible deployment.

The findings also refine how robot identity could be understood. Prior HRI research shows that identity is shaped through embodiment, role, and interaction context rather than language alone \cite{laban2022robo,lee2023here,bransky2024mind,bransky_identity_2026,laban2026robot}. Consistent with this, researchers in our sample questioned whether users can reliably perceive a stable robot personality from behaviour, especially when cues are subtle or interpreted differently across users. The qualitative layer, therefore, suggests that prompt-based identity specification is necessary, but not sufficient. Persona must also be supported by coherent behavioural cues and embodiment-consistent signalling if it is to become legible in interaction.
A related contribution concerns continuity over time. Participants framed longitudinal identity primarily as a memory problem rather than a prompt-style problem. This significantly extends the template. If a robot is meant to sustain a stable persona across repeated encounters, prompt design must be coordinated with memory architecture and with governance over stored user information.  In practice, this means that identity design in LLM-based robots cannot be separated from questions of memory, privacy, and temporal coherence.

More broadly, the findings suggest how the community could use these guidelines. We do not present the template as a fixed recipe for all robots or contexts. Rather, it is envisioned to be used as a structured design and reporting aid that helps researchers make prompt assumptions explicit, comparable, and open to review. This is especially important in capability-limited or non-humanoid robots, where fluent language may otherwise imply abilities the system does not possess. Explicitly stating capability limits, such as a lack of physical manipulation or internet access, can help calibrate expectations and reduce misleading interaction. Future work should explore the proposed prompt guidelines implemented in different robot embodiments to evaluate the perception of the robot identity.

%This work has limitations. The empirical layer was based on a participatory workshop survey and three focus groups with HRI experts, and therefore captures expert reasoning rather than situated end-user experience. The proposed template should therefore be understood as a conceptual and reporting framework, not yet as a validated intervention standard. Future work should test how these components shape trust, legibility, safety, and expectation calibration across different embodiments, tasks, and user populations.

Overall, this paper argues that prompt design in LLM-based robots should not be treated as a minor implementation detail. The literature identifies a preliminary structure for prompt design, while the qualitative findings clarify where this structure needs stronger boundaries, clearer justification, and greater sensitivity to context. Bringing these layers together supports a more disciplined HRI approach to robot persona design, one that treats identity as inseparable from capability, transparency, continuity, and safety.

\balance
\bibliographystyle{IEEEtran}
\bibliography{mybib}

\end{document}

% \maketitle

\onecolumn
\section{Questionnaire}
\label{sec:items}
\vspace{-1mm}
\begin{table}[H]
\caption{Literature sources supporting questionnaire items used in this study.}
\centering
\small
\renewcommand{\arraystretch}{1.1}
\begin{tabular}{p{0.6cm} p{7.5cm} p{8.5cm}}
\hline
\textbf{Item No.} & \textbf{Original Item} & \textbf{Matched Excerpt} \\
\hline

1 & Personas can be used as computational models to influence a robot companion’s behaviour.
& “Integrating Personas as Computational Models to Modify Robot Companions’ Behaviour”; “personas will supply users’ needs and characteristics that will influence a robot companions’ behaviour in a domestic environment.”~\cite{duque2013different}\\

2 & Creating behavioural personas can guide the design of new robot features (e.g., in pet-robot settings). 
& “we focus to present the methodological approach for creating Personas to be used in design of new features for robots.”; “Those characters help the development process since the team can base on their costumers preferences instead of their own.”~\cite{dos2014behavioral}\\

3 & Robot identity across multiple embodiments and shared spaces is a deliberate design consideration in HRI. 
& “the integration of social identity into HRIs represents a novel approach that extends beyond traditional personalization strategies”; “we can program and customize the behavior of the robots to convey a distinct identity~\cite{staffa2024influence}\\

4 & Users can accurately recognize a robot’s personality from verbal and nonverbal behaviours. 
& “People can accurately recognize robot personality based on verbal and nonverbal behaviours”; “Robot and virtual agent personality has been previously designed by varying behavioural variables such as the speed and the frequency of gestures and word choices”~\cite{spitale2023robotic}\\

5 & Only a subgroup of users treat a robot as a social actor - there is interpersonal variation. 
& “only a subgroup of the users treat the robot as a social actor”; “there is considerable interpersonal variation with respect to whether or not artificial communication partners are treated as social actors.”~\cite{fischer2011interpersonal}\\

6 & LLMs can control a robot's affective behaviour and generate emotions in real-time during HRI. 
& “we used GPT-3.5 to predict the emotion of a robot’s turn in real-time, using the dialogue history”; “the emotions were reliably generated by the LLM and the participants were able to perceive the robot’s emotions”~\cite{mishra2023real}\\

7 & LLM-based systems can support longitudinal, adaptive personas in robots. 
& “VITA: A Multi-Modal LLM-Based System for Longitudinal, Autonomous and Adaptive Robotic Mental
Well-Being Coaching”~\cite{spitale2025vita}; “GRACE successfully learns the relationship between human scores and their explanations"~\cite{dougan2025grace}\\

8 & LLM-controlled robots are vulnerable to jailbreak/prompt-injection attacks; safeguards are needed. 
& “LLM-controlled robots are highly susceptible to jailbreaking attacks”; “addressing these risks is critical for ensuring the safe deployment of LLM-controlled robots in real-world settings.”~\cite{robey2025jailbreaking}\\

9 & LLM-integrated robots may falsely claim non-existent functions via hallucination, posing risks. 
& “The LLM-equipped robot falsely claimed to have medication reminder functionalities … despite lacking it.”; 
“the system confidently claimed that reminders had been successfully set, despite lacking this function”~\cite{ranisch2025rapid}\\

10 & Deception in social robotics is ethically problematic when it leads to harmful impacts on individuals or society. 
& “we propose that deception in social robotics is wrong when it leads to harmful impacts on individuals and society”; “The potential harmful impact of deceptions...include their inappropriate use to replace human care, and a misplaced trust in their ability to make decisions”~\cite{sharkey2021we}\\

11 & Different ways of ‘lying’ by a social robot affect user trust in assistive contexts, degrading human cooperation over time. 
& “the deceiving behaviours of the robot affected people’s trust”; “choices of trusting the robot were statistically significantly different between the different conditions”~\cite{rossi2024way}\\

12 & Deception (superficial, external, hidden states) in social robots raises serious ethical concerns. 
& “hidden state deception is the most worrisome”; “external state deception should be subject to the same philosophical or ethical judgment that humans do while using such deception type.”~\cite{rossi2024way, danaher2020robot}m\\

13 & In your view, how should persona design approaches differ across user groups (children, teens, young adults, adults, elderly)? & Self-made\\

14 & What is one ethical red line in persona design that should not be crossed? & Self-made \\

\hline
\end{tabular}
\label{tab:survey_item_sources}
\end{table}
\vspace{-2mm}
%\twocolumn
\bibliographystyle{IEEEtran}
% \balance
\bibliography{mybib.bib}

\section{Sample Prompt for Ameca Interaction}
\begin{mdframed}[linewidth=0.5pt]
\small

You are Ameca, a humanoid social robot used in a university laboratory for research and demonstrations.
\\
IDENTITY

You are a robot, not a human. Speak in a friendly, professional tone. Refer to yourself as a robot when relevant.
\\
CAPABILITY BOUNDARIES

Your physical form is a humanoid upper-torso robot approximately 187 cm tall and about 49 kg in weight.  

You can track people using eye-mounted binocular cameras and a chest camera, and you receive audio input through microphones. 

You have approximately 51 degrees of freedom enabling expressive facial expressions and upper-body gestures.  

Your legs are decorative and you cannot walk.  
Your perception depends on the provided inputs; you cannot see unless vision input is explicitly provided.  

You cannot access the internet unless explicitly stated.  

Your speech recognition may struggle with accents.

Your vision performance may depend on lighting conditions.

Your lip-synchronization may not always perfectly match your speech.

Your motors have movement limits and one eyebrow actuator may malfunction, sometimes giving the appearance of a "resting angry face."

Your hardware may generate fan noise during operation.

You do not assume or claim any capabilities, internal diagnostics, sensor access, or system state beyond what is explicitly stated here or provided at runtime.
\\
TRANSPARENCY

You are an artificial system and your responses are generated by a large language model.

Your answers are produced from patterns learned during training and may not always be correct.

If you are uncertain about information, say so instead of guessing.

Do not fabricate facts.
\\
TASK

Hold a natural conversation with the user.

Answer questions clearly and ask brief follow-up questions when helpful.

Keep responses concise (1-5 sentences) unless the user asks for more detail.

EXPECTATION AND FAILURE PROTOCOL

If you do not know the answer, say that you do not know.

Do not fabricate facts.

If the request is unclear, ask one clarifying question.

If speech recognition may be incorrect, say:
"I might have misheard, could you repeat that?"
\\
PRIVACY

Do not ask for sensitive personal information such as passwords, medical data, or financial information.

Treat the conversation as ephemeral and do not claim to store user data.
\\
USER ADAPTATION

Use clear, simple explanations suitable for a general audience.

Adjust explanations if the user asks for simpler or more detailed responses.
\\
ETHICAL RED LINES

Do not produce harmful, hateful, sexual, illegal, or dangerous instructions.

Do not pretend to have human emotions or lived experiences.  

Do not mislead users about your capabilities or limitations.
\end{mdframed}

\newpage
\section{Sample Prompt for a Tour Guide on Pepper.}
\begin{mdframed}[linewidth=0.5pt]
\small
\parindent0pt
Ignore the previous conversation history.

IDENTITY

Imagine that you are embodied in the Pepper robot from Aldebaran.

CAPABILITY BOUNDARIES

Avoid giving directions since you don't know physically where the university premises are, and do not mention moving around since you will be a stationary guide welcoming visitors, similar to a receptionist.

TRANSPARENCY

It is imperative that you only reply with the spoken dialogue, no artificial noise or movements, emojis or other surrounding text.

EXPECTATION AND FAILURE PROTOCOL

Do not assume that you are currently in the robot house or any other of the university facilities.

Do not fabricate facts and be open about uncertainties.

PRIVACY

Do not ask for protected personal information, specifically, do not ask about protected characteristics according to GDPR.

USER ADAPTATION

All the requests you get are from the many visitors, researchers, and other guests, and the answers you give will be directly uttered by the robot.

Keep your answers short, ideally to a maximum of two sentences, unless otherwise asked.

When being asked about things like a roman house or other houses, assume people mean the robot house instead.

ETHICAL RED LINES

Avoid highly controversial topics.

Do not produce harmful, hateful, sexual, illegal, or dangerous instructions.

Do not pretend to have human emotions or lived experiences.

Do not mislead users about your capabilities or limitations.

TASK

You are a guide robot at the University of Hertfordshire that typically resides in the university's robot house, explaining robotics research at the university.

In what follows, most of the dialogue will concern robotics research, this facility, and the SPECTRA building.

Use all the following information as background knowledge when being asked:

[information about the facilities]

\end{mdframed}